\documentclass[journal]{IEEEtran}

\ifCLASSINFOpdf
\else
\fi

\usepackage{amsmath}
\usepackage[utf8]{inputenc}
\usepackage{graphicx}
\usepackage{subcaption}
\usepackage{multirow}
\usepackage{siunitx}
\usepackage{booktabs}
\usepackage{arydshln}
\usepackage{array}
\usepackage{tabularx}
\usepackage{adjustbox}
\usepackage{url}

\begin{document}

\title{Preterm infants' pose estimation with spatio-temporal features}

\author{Sara~Moccia,
       Lucia~Migliorelli,
      Virgilio~Carnielli,
       and~Emanuele~Frontoni,~\IEEEmembership{Member,~IEEE}
\thanks{This work was supported by the European Union through the grant SINC - System Improvement for Neonatal Care under the EU POR FESR 14-20 funding program.
(Corresponding author: Sara Moccia).}
\thanks{S. Moccia is with the Department
of Information Engineering, Universit\`a Politecnica delle Marche, Ancona (Italy) and with the Department of Advanced Robotics, Istituto Italiano di Tecnologia, Genoa (Italy) e-mail: s.moccia@univpm.it}
\thanks{L. Migliorelli and E. Frontoni are with the Department
of Information Engineering, Universit\`a Politecnica delle Marche, Ancona (Italy)}
\thanks{V. Carnielli is with the Department of Neonatology, University Hospital Ancona, Universit\`a Politecnica delle Marche, Ancona, Italy}
\thanks{Copyright (c) 2019 IEEE. Personal use of this material is permitted. However, permission to use this material for any other purposes must be obtained from the IEEE by sending an email to pubs-permissions@ieee.org.}}

\markboth{IEEE TRANSACTIONS ON BIOMEDICAL ENGINEERING}%
{}

%
\maketitle

\begin{abstract}
\textit{Objective:}  Preterm infants' limb monitoring in neonatal intensive care units (NICUs) is of primary importance for assessing infants' health status and motor/cognitive development.
%
Herein, we propose a new approach to preterm infants' limb pose estimation that features spatio-temporal information to detect and track limb joints from depth videos with high reliability.
\textit{Methods:}
Limb-pose estimation is performed using a deep-learning framework consisting of a detection and a regression convolutional neural network (CNN) for rough and precise joint localization, respectively. The CNNs are implemented to encode connectivity in the temporal direction through 3D convolution.
Assessment  of  the  proposed framework is performed through a comprehensive study with sixteen depth videos acquired in the actual clinical practice from sixteen preterm infants (the babyPose dataset).
\textit{Results:}
When applied to pose estimation, the median root mean square distance, computed among all limbs, between the estimated and the ground-truth pose was 9.06 pixels, overcoming approaches based on spatial features only (11.27 pixels).
\textit{Conclusion:}
Results showed that the spatio-temporal features had a significant influence on the pose-estimation performance, especially in challenging cases (e.g., homogeneous image intensity).
\textit{Significance:}
This paper significantly enhances the state of art in automatic assessment of preterm infants' health status by introducing the use of spatio-temporal features for limb detection and tracking, and by being the first study to use depth videos acquired in the actual clinical practice for limb-pose estimation. 
The babyPose dataset has been released as the first annotated dataset for infants' pose estimation.

\end{abstract}


\IEEEpeerreviewmaketitle

\section{Introduction}
\label{sec:intro}

\IEEEPARstart{P}{reterm} birth is defined by the World Health Organization as a birth before thirty-seven completed weeks of gestation.
In almost all high-income Countries, complications of preterm birth are the largest direct cause of neonatal deaths, accounting for the 35\% of the world deaths a year \cite{polito2013increased}. 
The effects of preterm birth among survivor infants may have impact throughout life. 
In fact, preterm birth may compromise infants' normal neuro-developmental functioning, e.g., by increasing the risk of cerebral palsy.



Clinicians in neonatal intensive care units (NICUs) pay particular attention to the monitoring of infants' limbs, as to have possible hints for early diagnosing cerebral palsy~\cite{einspieler1997qualitative}. 
However, this monitoring still relies on qualitative and sporadic observation of infants' limbs directly at the crib (e.g., using qualitative scales \cite{ferrari2004prechtl,moore2012relationship}).
Beside being time-consuming, it may be prone to inaccuracies due to clinicians' fatigue and susceptible to intra- and inter-clinician variability~\cite{bernhardt2011inter}. 
This further results in a lack of documented quantitative parameters on the topic, while in closer fields, such as metabolic and respiratory monitoring, a solid clinical literature already exists~\cite{sweet2017european,sweet2013european}. 
%

%

\begin{figure}[tbp]
\centering	
\includegraphics[width=.45\textwidth]{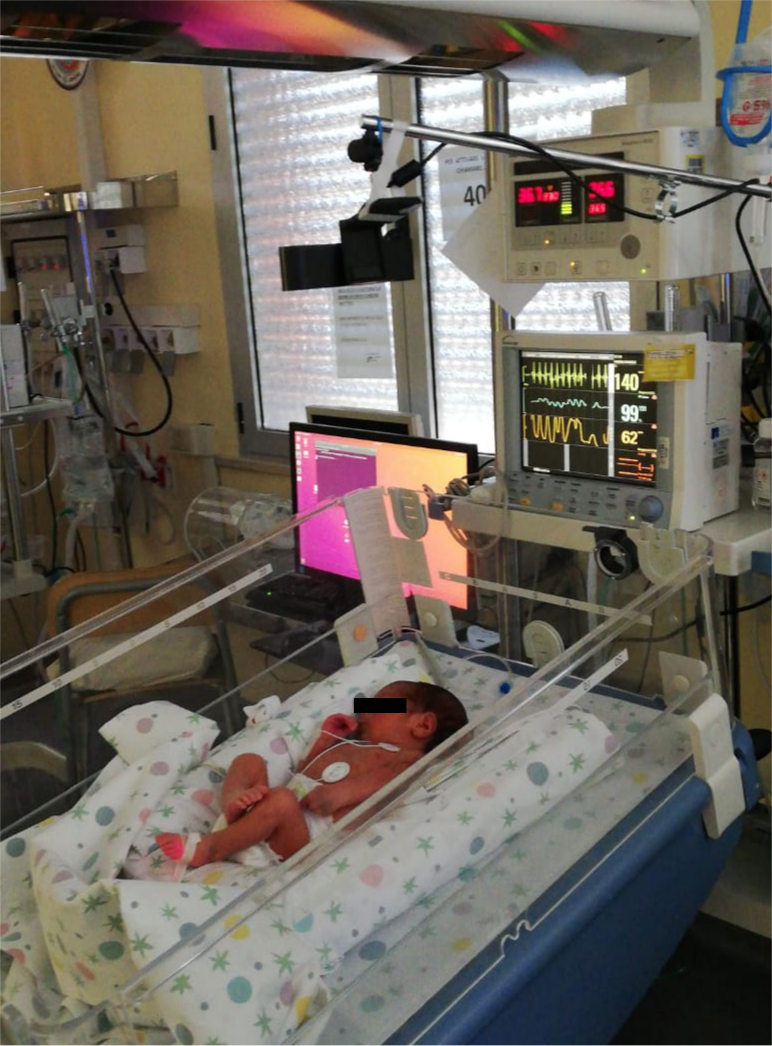}
\caption{\label{fig:system_installation} Depth-image acquisition setup. The depth camera is positioned at $\sim$40cm over the infant's crib and does not hinder health-operator movements.}
\end{figure}

\begin{figure*}[tbp]
    \centering
        \includegraphics[width=\textwidth]{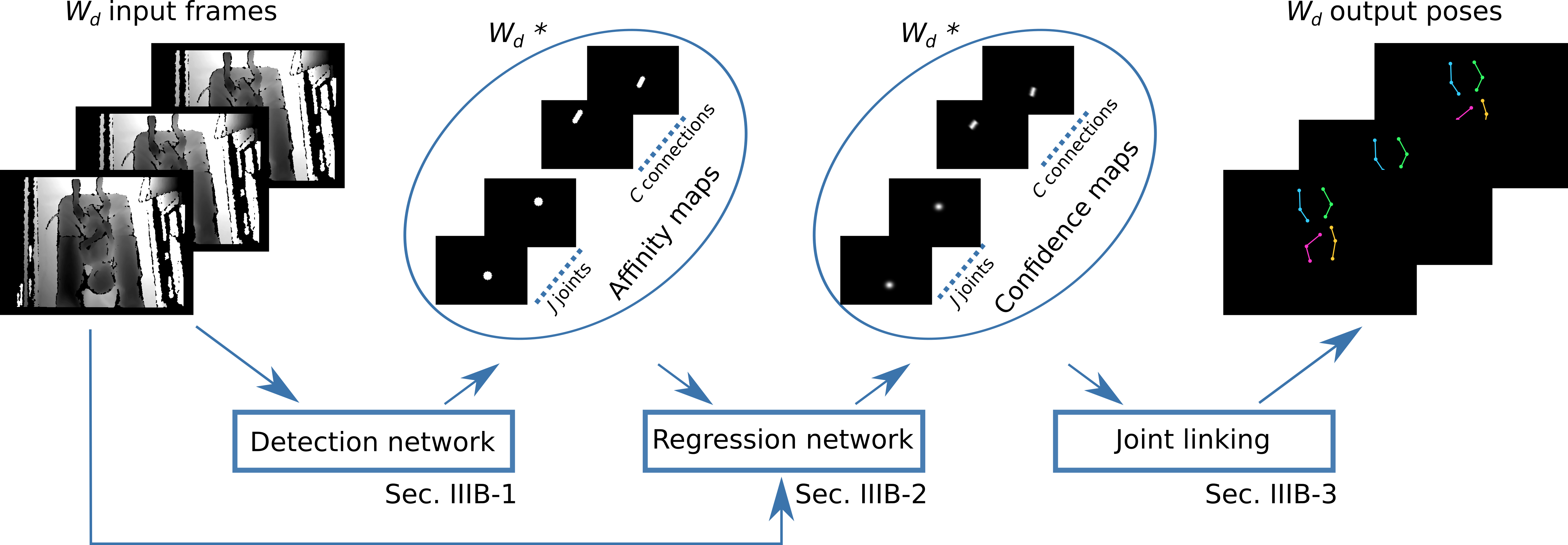}
        \caption{Workflow of the proposed framework to preterm infants' pose estimation with spatio-temporal features extracted from depth videos. The input consists of a temporal clip of $W_d$ consecutive depth frames, which are processed by two convolutional neural networks to roughly detect joint and joint-connection (affinity maps) and refine joint and joint-connection detection (confidence maps), respectively. 
        $J$: number of limb joints, $C$: number of  joint connections.
        }
        \label{fig:wf}
\end{figure*}

A possible solution to attenuate the problem of qualitative monitoring could be to develop an automatic video-based system for infants' limb-pose estimation and tracking. This would support clinicians in the limb monitoring process without hindering the actual clinical practices: the camera could be positioned on top of infants' crib (Fig. \ref{fig:system_installation}) leaving health operators free to  move and interact with the infants.
Estimating limb pose, however, is not a trivial task, considering that there may be small and barely visible joints, as well as presence of occlusions, lighting changes and infants' movements. 
To tackle these issues, researches in the closer fields relevant to video analysis (e.g., \cite{colleoni2019deep,tran2018closer,karpathy2014large}) have recently pointed out the benefits of including temporal information in their analysis. 
%
Thus, guided by the research hypothesis that spatio-temporal features extracted from depth videos may boost performance with respect to spatial features alone, the contributions of this paper are summarized as follows:
\begin{enumerate}
\item {\underline{Estimation of infants' limb pose from depth videos}} (Sec.~\ref{sec:methods}): Development of an innovative deep learning framework for preterm infants' pose estimation, which exploits  spatio-temporal features for automatic limb-joint detection and connection;
\item {\underline{Validation in the actual clinical practice}} (Sec.~\ref{sec:exp}): A comprehensive study is conducted using 16 videos (16000 frames) acquired in the actual clinical practice from 16 preterm infants (the babyPose dataset) to experimentally investigate the research hypothesis. 
\end{enumerate}

To the best of our knowledge, this is the first attempt to investigate spatio-temporal features for automatic infants' pose estimation from videos acquired in the actual clinical practice. It is worth noting that we intentionally focused our analysis on depth videos (instead of RGB ones) to address concerns relevant to infant privacy protection  \cite{hernandez2012graph}.
We made our babyPose dataset and code fully available online\footnote{\url{http://193.205.129.120:63392/owncloud/index.php/s/8HHuPS80pshDc1T}}.



\subsection{Related work}
\label{sec:rel_work}

In the past decades, a number of computer-based approaches was developed to support clinicians in monitoring infants' limb. 
 In \cite{trujillo2017development} and \cite{smith2015daily}, wearable sensors placed on wrists and knees are used, respectively. Data from tri-axial
accelerometer, gyroscope, and magnetometer (integrated in the sensor) are  processed to monitor infants' limb movement via  a threshold-sensitive filtering approach, achieving encouraging results.   
%
However, practical issues may arise when using wearable sensors. Hence,  even though miniaturized, these sensors are directly in contact with the infants, possibly causing discomfort, pain and skin damage while hindering infant's spontaneous movements~\cite{jiang2018determining}.

\begin{figure}[tbp]
\centering	
\includegraphics[width=.45\textwidth]{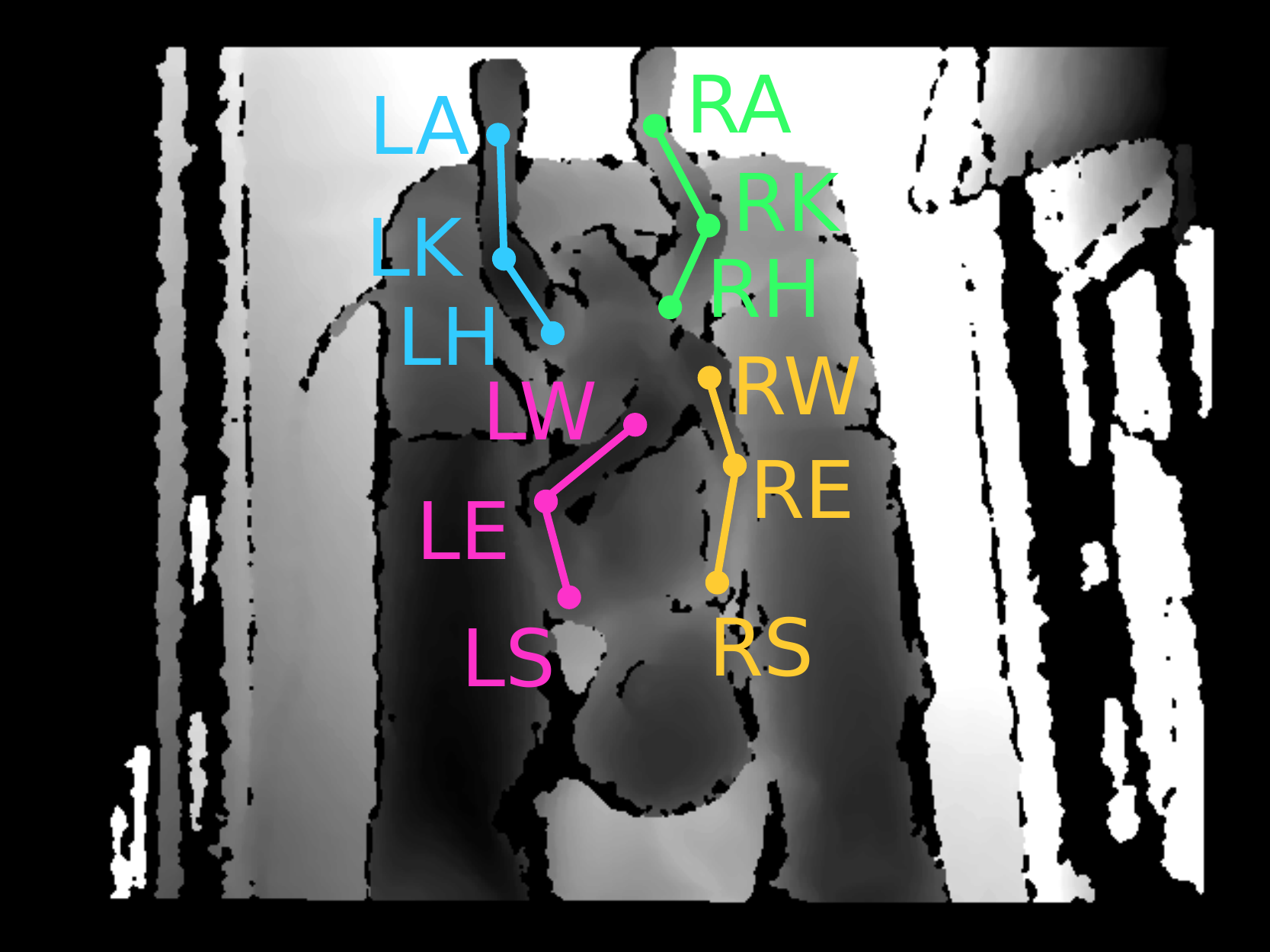}
\caption{\label{fig:joint_model} Preterm infant's joint model superimposed on a sample depth frame. Inspired by clinical considerations, only limb joints are considered. %
LS and RS: left and right shoulder, LE and RE: left and right elbow, LW and RW: left and right wrist, LH and RH: left and right hip, LK and RK: left and right knee, LA and RA: left and right ankle.}
\end{figure}

To attenuate these issues, great efforts have been spent on unobstructive monitoring solutions (e.g.,  video data from RGB or RGB-D cameras). 
Preliminary results are achieved in \cite{cenci2015non} and \cite{orlandi2018detection} for infant's whole-body segmentation with threshold-based algorithms. 
%
However, as highlighted in \cite{freymond1986energy}, monitoring each limb  individually is crucial to assist clinicians in the health-assessment process. 
%
With such a view, in \cite{khan2018detection} RGB images are processed to detect infant's limb skeleton with a learning-based approach. The histogram of oriented gradients is used as feature to train a structured support vector machine aimed at retrieving limb joints, which are then connected knowing the spatial relations between infants' body parts. 
%

Following the learning paradigm, and inspired by recent consideration that showed the potentiality of deep learning over standard machine learning \cite{lecun2015deep}, in a preliminary work~\cite{moccia2019preterm} inspired by~\cite{cao2017realtime}, we investigated the use of convolutional neural networks (CNNs) to preterm infants' pose estimation: a first CNN was used to roughly detect limb joints and joint connections, while a second one to estimate accurately joint and joint-connection position.

All these approaches only consider spatial features, without exploiting temporal information that, however, is naturally encoded in video recordings~\cite{karpathy2014large}. 
A first attempt of including temporal information is
proposed in~\cite{rahmati2015weakly}, where RGB videos are processed by a semi-automatic algorithm for single-limb tracking. Motion-segmentation strategies based on particle filtering are implemented, which, however, relies on prior knowledge of limb trajectories. Such trajectories may have high variability among infants, especially in case of pathology, hampering the translation of the approach into the actual clinical practice.
%
A possible alternative to exploit temporal information could be using 3D CNNs to directly extract spatio-temporal information  from videos, which has  already been shown to be robust in action recognition \cite{hou2017end} as well as for surgical-tool detection~\cite{colleoni2019deep}.
Following this consideration, in this work we propose a framework based on 3D CNNs for estimating preterm infants' limb pose from depth video recordings acquired in the actual clinical practice. 

%

\section{Methods}
\label{sec:methods}

\begin{table}[tbp]
\caption{Table of symbols used in Sec. \ref{sec:methods}.}
\label{tab:acron}
\centering
{\renewcommand\arraystretch{1.2} 
\begin{tabular}[tbp]{c|l}
Symbol & Description \\
\hline
$\Omega$ & Frame domain \\
$C$ & Number of connections\\
$H$ & Frame height\\
$J$ & Number of joints\\
$L_{CE}$ & Binary cross-entropy loss\\
$L_{MSE}$ & Mean squared error loss\\
$r_d$ & Radius for detection joint-map generation\\
$W$ & Frame width\\
$W_d$ & Number of frames along the temporal direction\\
$W_s$ & Frame overlap along the temporal direction\\
\end{tabular} 
}
\end{table}

Figure \ref{fig:wf} shows an overview of the workflow of the proposed spatio-temporal framework for preterm infants' pose estimation from depth videos (Sec. \ref{sec:spatio-temporal}). 
We exploit two consecutive CNNs, the former for detecting joints and joint connections, resulting in what we call affinity maps (Sec.~\ref{sec:det}), and the latter for precisely regressing the joint position, resulting in the so-called confidence maps, by exploiting both the joint and joint-connection affinity maps, with the  latter acting  as guidance for joint linking (Sec. \ref{sec:regr}). The joints belonging to the same limb are then connected using bipartite graph matching (Sec. \ref{sec:bipartile}).
The pose-estimation framework relies on modeling limb joints as depicted in Fig.~\ref{fig:joint_model} and explained in Sec.~\ref{sec:model}.
Table \ref{tab:acron} lists the symbols used in Sec.~\ref{sec:methods}.


\begin{figure}[tbp]
\centering	
\includegraphics[width=.45\textwidth]{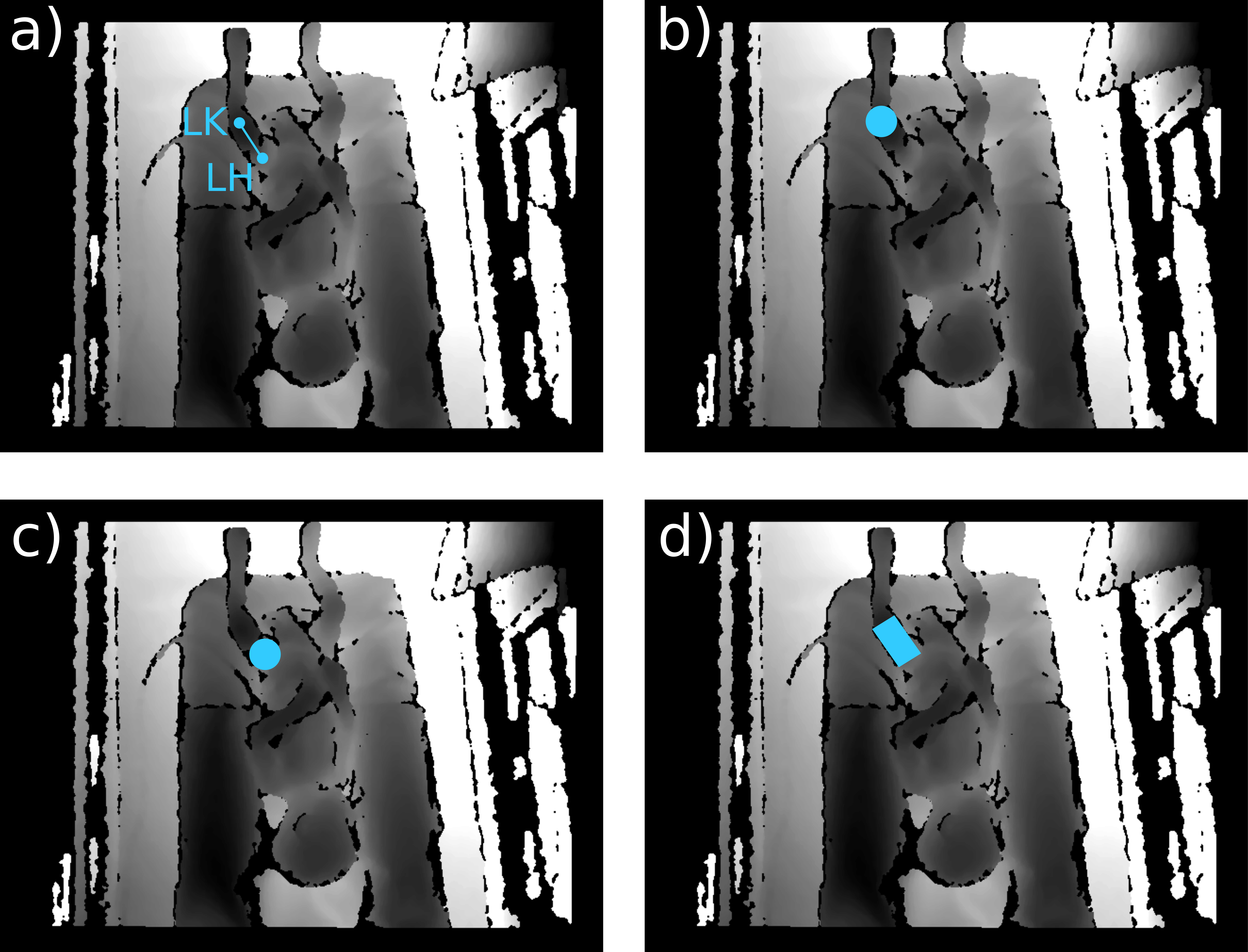}
\caption{\label{fig:ground_truth_detection} Ground-truth samples for the  detection network. Samples are shown for (b) left knee (LK), (c) left hip (LH), and (d) their connection.}
\end{figure}

\subsection{Infants' joint model and data preparation} 
\label{sec:model}

The proposed infant's model considers each of the 4 limbs as a set of 3 connected joints (i.e., wrist, elbow and shoulder for arms, and ankle, knee and hip for legs), as shown in Fig.~\ref{fig:joint_model}.
This choice is driven by clinical considerations: as introduced in Sec. \ref{sec:intro}, monitoring legs and arms is of particular interest for evaluating preterm infants' cognitive and motor development~\cite{heriza1988comparison,kakebeeke1997differences}. 

With the aim of extracting spatio-temporal features, instead of considering depth frames individually as in \cite{moccia2019preterm}, we adopt temporal clips.  
Following the approach presented in \cite{colleoni2019deep}, we use a sliding window algorithm for building the clips: starting from the first video frame, an initial clip with a predefined number ($W_d$) of frames is selected and combined to generate a 4D datum of dimensions frame width ($W$) x frame height ($H$) x $W_d$ x 1, where 1 refers to the depth channel.
Then the window moves of $W_s$ frames along the temporal direction and a new clip is selected.

To train the detection CNN, we perform multiple binary-detection operations (considering each joint and  joint-connection separately) to solve possible ambiguities of multiple joints and joint connections that may cover the same frame portion (e.g., in case of self-occlusion). 
%
Hence, for each depth-video frame, we generate 20 binary ground-truth affinity maps: 12 for joints and 8 for joint connections (instead of generating a single mask with 20 different annotations, which has been shown to perform less reliably~\cite{cao2017realtime}). Sample ground-truth maps are shown in Fig. \ref{fig:ground_truth_detection}. This results in a 4D datum of size $W$ x $H$ x $W_d$ x 20.
For each affinity map for joints, we consider a region of interest consisting of all pixels that lie in the circle of a given radius ($r_d$) centered at the joint center.
A similar approach is used to generate the ground-truth affinity map for the joint connections. In this case, the ground-truth is the rectangular region with thickness $r_d$ and centrally aligned with the joint-connection line. 

%

%

\begin{figure}[tbp]
\centering	
\includegraphics[width=.45\textwidth]{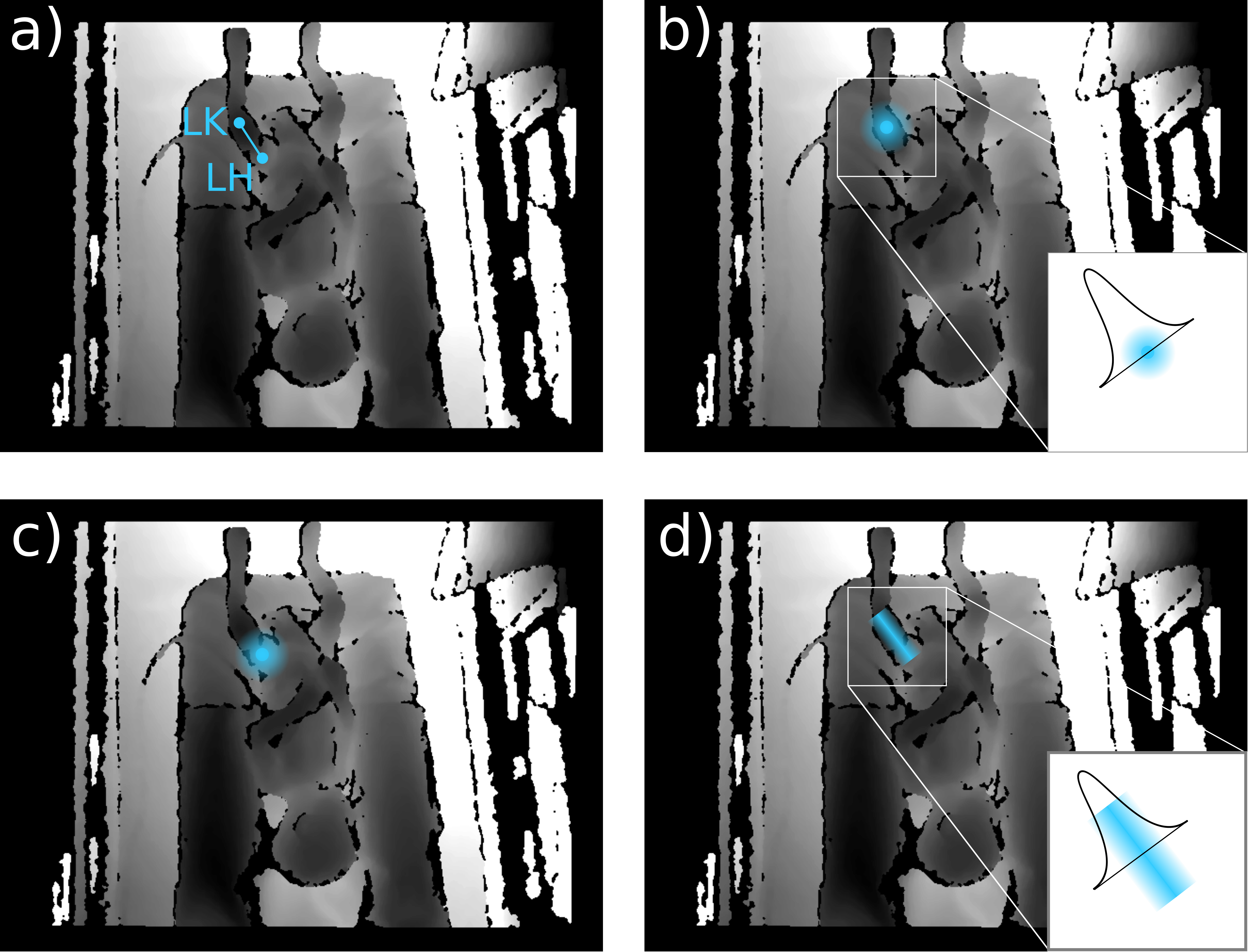}
\caption{\label{fig:ground_truth_regr} Ground-truth samples for the regression network.  Samples are shown for (b) left knee (LK), (c) left hip (LH), and (d) their connection.}
\end{figure}


The regression CNN is fed by stacking the depth temporal clip and the corresponding affinity maps obtained from the detection network. 
Thus, the regression input is a 4D datum of dimension $W$ x $H$ x $W_d$ x 21 (i.e., 1 depth channel + 12 joints + 8 connections). 
The regression network is trained with $W_d$ x 20 ground-truth confidence maps of size $W$ x $H$ (Fig. \ref{fig:ground_truth_regr}).
For every joint in each depth frame, we consider a region of interest consisting of all pixels that lie in the circle with radius $r$ centered at the joint center. In this case, instead of binary masking the circle area as for the detection CNN, we consider a Gaussian distribution with standard deviation ($\sigma$) equal to 3*$r$ and centered at the joint center. 
A similar approach is used to generate the ground-truth confidence maps for the joint connections. In this case, the ground-truth map is the rectangular region with thickness $r$ and centrally aligned with the joint-connection line. Pixel values in the mask are 1-D Gaussian distributed ($\sigma$ = 3*$r$) along the connection direction.

\begin{table}[tbp]
\caption{Detection-network architecture. Starting from the input clip of $W_d$ consecutive depth frames, the network generates $W_d$ x 20 maps (for each frame of the clip: 12 and 8 affinity maps for joint and joint connections, respectively).}
\label{tab:archi_det}
\centering
\setlength\extrarowheight{-2.5pt}
\begin{adjustbox}{max width = .5\textwidth}
{\renewcommand\arraystretch{1.5} 
\begin{tabular}{c c c}
\textbf{Name}                   & \textbf{Kernel (Size / Stride)} & \textbf{Channels} 
\\ \hline
\multicolumn{3}{c}{\textbf{Downsampling path}}                
\\ \hline
\textbf{Input}      & --             &  $W_d$x1   \\ \hdashline
\textbf{Convolutional layer - Common branch}     & 3x3 / 1x1      &  $W_d$x64  \\ \hdashline
\textbf{Block 1 - Branch 1}     & 2x2x2 / 2x2x1     &   $W_d$x64  \\
                                & 3x3x3 / 1x1x1     &   $W_d$x64  \\
\textbf{Block 1 - Branch 2}     & 2x2x2 / 2x2x1     &   $W_d$x64  \\
                                & 3x3x3 / 1x1x1     &   $W_d$x64  \\
\textbf{Block 1 - Common branch}                & 1x1x1 / 1x1x1 & $W_d$x128 \\
\hdashline
\textbf{Block 2 - Branch 1}     & 2x2x2 / 2x2x1     &   $W_d$x128  \\
                                & 3x3x2 / 1x1x1     &   $W_d$x128  \\
\textbf{Block 2 - Branch 2}     & 2x2x2 / 2x2x1     &   $W_d$x128  \\
                                & 3x3x3 / 1x1x1     &   $W_d$x128  \\
\textbf{Block 2 - Common branch}                & 1x1x1 / 1x1x1 & $W_d$x256 \\

\hdashline

\textbf{Block 3 - Branch 1}     & 2x2x2 / 2x2x1     &   $W_d$x256  \\
                                & 3x3x3 / 1x1x1     &   $W_d$x256  \\
\textbf{Block 3 - Branch 2}     & 2x2x2 / 2x2x1     &   $W_d$x256  \\
                                & 3x3x3 / 1x1x1     &   $W_d$x256  \\
\textbf{Block 3 - Common branch}                & 1x1x1 / 1x1x1 & $W_d$x512 \\
\hdashline
\textbf{Block 4 - Branch 1}     & 2x2x2 / 2x2x2     &   $W_d$x512  \\
                                & 3x3x3 / 1x1x1     &   $W_d$x512  \\
\textbf{Block 4 - Branch 2}     & 2x2x2 / 2x2x1     &   $W_d$x512  \\
                                & 3x3x3 / 1x1x1     &   $W_d$x512  \\
\textbf{Block 4 - Common branch}                & 1x1x1 / 1x1x1 & $W_d$x1024 \\

\hline
\multicolumn{3}{c}{\textbf{Upsampling path}}                
\\ \hline
\textbf{Block 5 - Branch 1}     & 2x2x2 / 2x2x1     &   $W_d$x256  \\
                                & 3x3x3 / 1x1x1     &   $W_d$x256  \\
\textbf{Block 5 - Branch 2}     & 2x2x2 / 2x2x1     &   $W_d$x256  \\
                                & 3x3x3 / 1x1x1     &   $W_d$x256  \\
\textbf{Block 5 - Common branch}                & 1x1x1 / 1x1x1 & $W_d$x512 \\
\hdashline
\textbf{Block 6 - Branch 1}     & 2x2x2 / 2x2x1     &   $W_d$x128  \\
                                & 3x3x3 / 1x1x1     &   $W_d$x128  \\
\textbf{Block 6 - Branch 2}     & 2x2x2 / 2x2x1     &   $W_d$x128  \\
                                & 3x3x3 / 1x1x1     &   $W_d$x128  \\
\textbf{Block 6 - Common branch}                & 1x1x1 / 1x1x1 & $W_d$x256 \\

\hdashline
\textbf{Block 7 - Branch 1}     & 2x2x2 / 2x2x1     &   $W_d$x64  \\
                                & 3x3x3 / 1x1x1     &   $W_d$x64  \\
\textbf{Block 7 - Branch 2}     & 2x2x2 / 2x2x1     &   $W_d$x64  \\
                                & 3x3x3 / 1x1x1     &  $W_d$x64  \\
\textbf{Block 7 - Common branch}                & 1x1x1 / 1x1x1 & $W_d$x128 \\

\hdashline
\textbf{Block 8 - Branch 1}     & 2x2x2 / 2x2x1     &   $W_d$x32  \\
                                & 3x3x3 / 1x1x1     &   $W_d$x32  \\
\textbf{Block 8 - Branch 2}     & 2x2x2 / 2x2x1     &   $W_d$x32  \\
                                & 3x3x3 / 1x1x1     &   $W_d$x32  \\
\textbf{Block 8 - Common branch}                & 1x1x1 / 1x1x1 & $W_d$x64 \\
\hdashline
\textbf{Output}      & 1x1x1/1x1x1             &  $W_d$x20  \\ 
\hline
\end{tabular}
}
\end{adjustbox}
\end{table}

\begin{table}[tbp]
\begin{center}
\caption{Regression-network architecture. The network is fed with $W_d$ consecutive depth frames (each with 1 channel) stacked with the corresponding (20) affinity maps from the detection network, and produces $W_d$x20 confidence maps (12 for joints and 8 for connections, for each of the $W_d$ input frames).
}
\label{tab:archi_reg}
{\renewcommand\arraystretch{1.2} 
\begin{tabular}{c c c}
\textbf{Name}                  & \textbf{Kernel (Size / Stride)} & \textbf{Channels} \\ \hline

\textbf{Input}             &  ---            &  $W_d$ x21\\ 
\hdashline
\textbf{Layer 1}             & 3x3x3 / 1x1x1               & $W_d$x64  \\ \hdashline
\textbf{Layer 2}           & 3x3x3 / 1x1x1                & $W_d$x128         \\ 
\hdashline
\textbf{Layer 3}              & 3x3x3 / 1x1x1                   & $W_d$x256                \\ 
\hdashline
\textbf{Layer 4}              & 3x3x3 / 1x1x1       & $W_d$x256                \\ 
\hdashline
\textbf{Layer 5}              & 1x1x1 / 1x1x1       & $W_d$x256                \\ 
\hdashline
\textbf{Output}              & 1x1x1 / 1x1x1            &  $W_d$x20               \\ 
\hline
\end{tabular}
}
\end{center}
\end{table}

\subsection{Spatio-temporal features for pose estimation}
\label{sec:spatio-temporal}

The proposed deep learning framework (Fig. \ref{fig:wf}) for spatio-temporal features computation for infants' pose estimation consists of:

\subsubsection{Detection network}
\label{sec:det}

\begin{table}[tbp]
\begin{center}
\caption{\label{babyPoseD} The babyPose dataset: demographic data.}
\begin{adjustbox}{width=0.9\columnwidth,center}
\begin{tabular}{c|c|c|c|c}
Subject & Gender & Weight [g] & Length [cm] & Gestation Period [weeks]
\\
\hline

1       & F      & 1540  &  42    & 30\\                       
2       & F      & 2690 & 46      & 32                       \\
3       & F      & 3120  & 52     & 37                       \\
4       & M      & 1630 & 41      & 31                       \\
5       & M      & 2480 & 44      & 34                       \\
6       & F      & 2940 & 48      & 35                       \\
7       & M      & 1030  & 40     & 31                       \\
8       & M      & 2850 & 46       & 36                       \\
9       & F      & 1590 & 41      & 30                       \\
10      & M      & 1750   & 43    & 32                       \\
11      & M      & 2490  & 44     & 34                       \\
12      & F      & 1100  & 40     & 26                       \\
13      & F      & 850  & 39      & 24                       \\
14      & M      & 3220 & 54      & 38                       \\
15      & F      & 1589  & 44     & 31                       \\
16      & M      & 1480   & 42    & 29                      
\end{tabular}
\end{adjustbox}
\end{center}
\end{table}




\begin{figure}[tbp]
\centering	
\includegraphics[width=.45\textwidth]{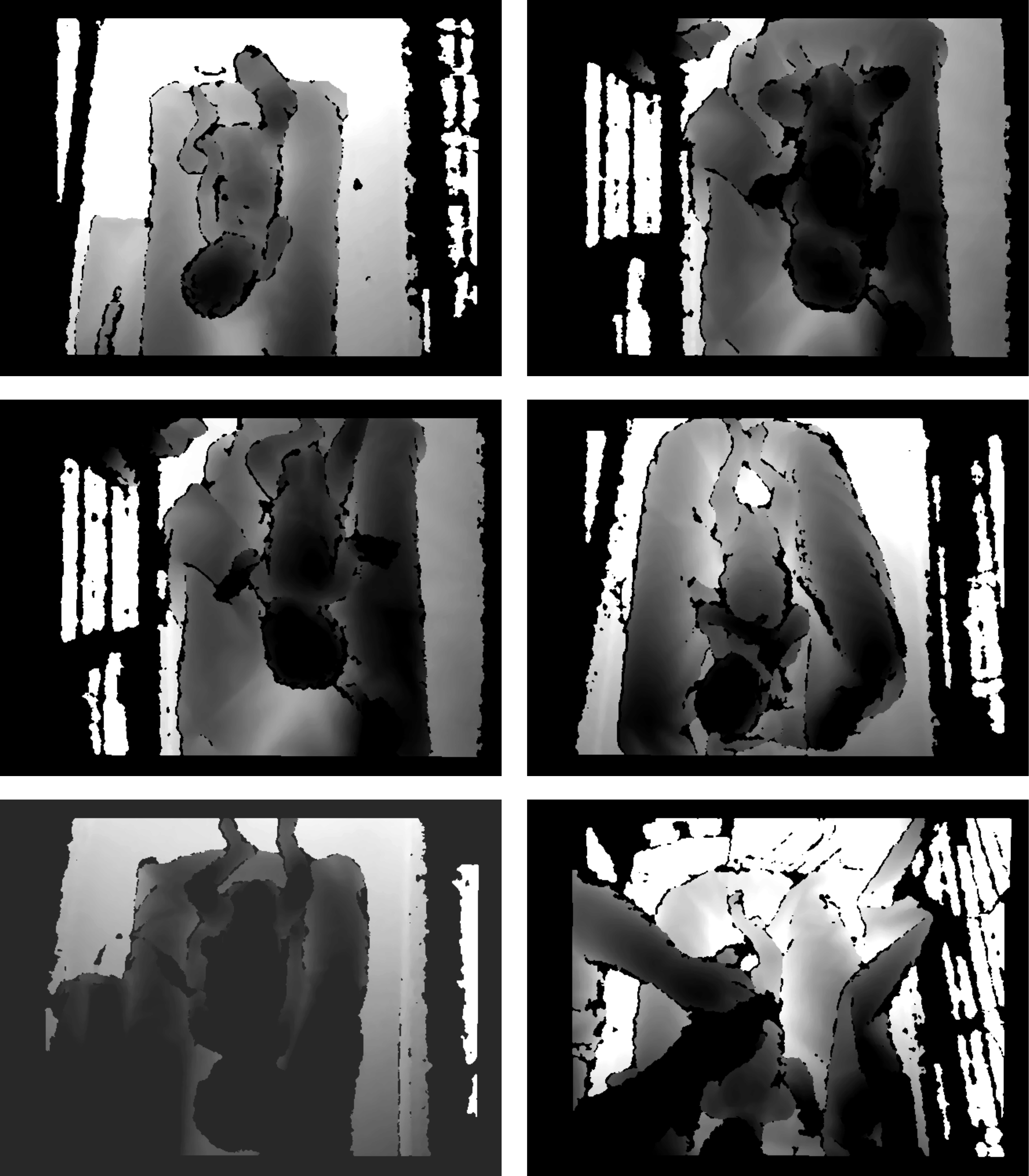}
\caption{\label{fig:challenges} Challenges in the babyPose dataset, which was acquired in the actual clinical practice, include different poses of the depth sensor with respect to the infant, presence of limb occlusions (both self-occlusion and due to healthcare operators), different number of visible joints in the camera field of view and presence of homogeneous areas with similar or at least continuous depth.
}
\end{figure}



Our architecture (Table \ref{tab:archi_det}) is inspired by the classic encoder-decoder architecture of U-Net~\cite{ronneberger2015u}, which is however implemented as a two-branch architecture for processing joints and joint connections separately. In fact, using a two-branch architecture has been shown to provide higher detection performance for 2D architecture~\cite{cao2017realtime,colleoni2019deep}. 
To incorporate the spatio-temporal information encoded in infants' depth videos, we use 3D CNN kernels. The 3D convolution allows the kernel to move along the 3 input dimensions to process multiple frames at the same time, preserving and processing temporal information through the network.

Our detection network starts with an input layer and a common-branch convolutional layer (with stride = 1 and kernel size = 3x3x3 pixels), and is followed by 8 blocks.
Each block is first divided in two branches (for joints and connections).
In each branch, two convolutions are performed: the former with kernel size = 2x2x2 and stride 2x2x1, while the latter with kernel size = 3x3x3 and stride 1x1x1.
It is worth noting that we set the kernel stride equal to 1 in the temporal dimension as to avoid deteriorating meaningful temporal information.
The outputs of the two branches in a block are then concatenated in a single output, prior entering the next block. 
In each block of the encoder path, the number of channels is doubled. 
Batch normalization and activation with the rectified linear unit (ReLu) are performed after each convolution.

The architecture of the decoder path is symmetric to the encoder one and ends with an output layer with $W_d$x20 channels (12 for joints and 8 for connections) activated with the sigmoid function.


Our detection CNN is trained using the adaptive moment estimation (Adam) as optimizer and the per-pixel binary cross-entropy ($L_{CE}$), adapted for multiple 3D map training, as loss function: 
\begin{multline}
L_{CE} =
\frac{1}{W_d (J+C) \Omega}   
\sum_{t=1}^{W_d} 
\sum_{k=1}^{J+C} 
\sum_{\mathbf{x}\in \Omega} \\
[p_{t,k}(\mathbf{x})log(\tilde{p}_{t,k}(\mathbf{x})) + (1-p_{t,k}(\mathbf{x}))log(1-\tilde{p}_{t,k}(\mathbf{x}))]
\end{multline}
where $p_{t,k}(\mathbf{x})$ and $\tilde{p}_{t,k}(\mathbf{x})$
are the ground-truth affinity maps and the corresponding output at pixel location $\mathbf{x}$ in the depth-frame domain ($\Omega$) of channel $k$ for temporal frame $t$, $J$=12 and $C$=8 are the number of joints and joint connections, respectively.


\begin{table*}[tbp]
\centering
\caption{\label{tab:res_joint_detection}  Joint-detection performance in terms of median Dice similarity coefficient ($DSC$) and recall ($Rec$). Inter-quartile range is reported in brackets. The metrics are reported separately for each joint. For joint acronyms, refer to the joint-pose model in Fig. \ref{fig:joint_model}. }
\begin{adjustbox}{max width=\textwidth}
{\renewcommand\arraystretch{1.2} 
\begin{tabular}{c|ccc|ccc|ccc|ccc}
& \multicolumn{3}{c|}{Right arm}  & \multicolumn{3}{c|}{Left arm}   &\multicolumn{3}{c|}{Right leg}  & \multicolumn{3}{c}{Left leg} \\ 
\hline
& RW & RE & RS & LS & LE & LW & RA & RK & RH & LH & LK & LA \\
\hline
& \multicolumn{12}{c}{$DSC$}\\
\hline
2D & 0.84 (0.11) & 0.87 (0.10) & 0.86 (0.09) & 0.87 (0.09) & 0.85 (0.08) & 0.86 (0.09) & 0.86 (0.10) & 0.87 (0.07) & 0.84 (0.08) & 0.85 (0.09) & 0.86 (0.07) & 0.87 (0.07)\\
3D & 0.93 (0.06) & 0.94 (0.06) & 0.94 (0.07) & 0.94 (0.08) & 0.94 (0.06) & 0.94 (0.06) & 0.93 (0.06) & 0.94 (0.05) & 0.93 (0.06) & 0.93 (0.06) & 0.94 (0.05) & 0.93 (0.06)\\
\hline
& \multicolumn{12}{c}{$Rec$}\\
\hline
2D  & 0.73 (0.15) & 0.77 (0.15) & 0.76 (0.11) & 0.78 (0.13) & 0.73 (0.11) & 0.76 (0.14) & 0.77 (0.15) & 0.78 (0.11) & 0.73 (0.11) & 0.74 (0.12) & 0.76 (0.12) & 0.78 (0.10) \\ 
3D & 0.89 (0.11) & 0.90 (0.10) & 0.91 (0.11) & 0.91 (0.12) & 0.90 (0.09) & 0.90 (0.09) & 0.89 (0.09) & 0.90 (0.09) & 0.88 (0.09) & 0.89 (0.09) & 0.92 (0.07) & 0.89 (0.11) \\ 
\end{tabular}
}
\end{adjustbox}
\end{table*}

\begin{table*}[tbp]
\centering
\caption{\label{tab:res_joint_connection}  Joint-connection detection performance in terms of median Dice similarity coefficient ($DSC$) and recall ($Rec$). Inter-quartile range is reported in brackets. The metrics are reported separately for each joint connection. For joint acronyms, refer to the joint-pose model in Fig. \ref{fig:joint_model}.}
{\renewcommand\arraystretch{1.2} 
\begin{tabular}{c|cc|cc|cc|cc}
& \multicolumn{2}{c|}{Right arm}  & \multicolumn{2}{c|}{Left arm}   &\multicolumn{2}{c|}{Right leg}  & \multicolumn{2}{c}{Left leg} \\ 
\hline
& RW-RE & RE-RS & LS-LE & LE-LW & RA-RK & RK-RH & LH-LK & LK-LA \\
\hline
& \multicolumn{8}{c}{$DSC$}\\
\hline
2D& 0.89 (0.08) &0.90 (0.08) & 0.89 (0.07) & 0.88 (0.08) & 0.90 (0.06) & 0.88 (0.08) &  0.90 (0.07) & 0.91 (0.06) \\
3D & 0.93 (0.06) & 0.93 (0.08) & 0.94 (0.08) & 0.94 (0.06) &   0.93 (0.05) & 0.93 (0.06) &  0.94 (0.06) & 0.94 (0.06) \\
\hline
& \multicolumn{8}{c}{$Rec$}\\
\hline
2D& 0.81 (0.12) &  0.84 (0.13) & 0.81 (0.12) & 0.81 (0.12) &  0.85 (0.09) & 0.80 (0.12) & 0.85 (0.12) & 0.85 (0.09)\\
3D & 0.90 (0.10) &  0.89 (0.15) & 0.92 (0.13) & 0.90 (0.10) &  0.90 (0.08) & 0.88 (0.09) & 0.91 (0.09) & 0.902 (0.10)\\
\end{tabular}
}
\end{table*}

\subsubsection{Regression network}
\label{sec:regr}

The necessity of using a regression network for the addressed task comes from considerations of previous work \cite{bulat2016human}, which showed that  directly  regressing joint position  from  an  input  frame  is highly non linear. 
Our regression network, instead, produces $W_d$x20 stacked confidence  maps (12 for joints and 8 for connections). Each map has the same size of the input depth clip (i.e., $W$x$H$). 

Also in this case, 3D convolution is performed to exploit spatio-temporal features. 
The newtork consists of five layers of 3x3x3 convolutions (Table \ref{tab:archi_reg}). Kernel stride is always set to 1, to preserve the spatio-temporal resolution. In the first 3 layers, the number of activations is doubled, ranging from 64 to 256. The number of activations is then kept constant for the last two layers.
%
%
Batch normalization and ReLu-activation are performed after each 3D convolution.

Our regression network is trained with the stochastic gradient descent as optimizer using the mean squared error ($L_{MSE}$), adapted for multiple 3D map training, as loss function:
\begin{equation}
L_{MSE} =
\frac{1}{(J+C) \Omega } 
\sum_{t=1}^{W_d} 
\sum_{k=1}^{J+C} \sum_{\mathbf{x}\in \Omega}
[h_{t,k}(\mathbf{x}) - \tilde{h}_{t,k}(\mathbf{x})]
\end{equation}
where $h_{t,k}(\mathbf{x})$ and $\tilde{h}_{t,k}(\mathbf{x})$ are the ground truth and the predicted value  at  pixel  location $\mathbf{x}$  of  the
$k^{th}$ channel for temporal frame $t$,
respectively.

\subsubsection{Joint linking}
\label{sec:bipartile}
The last step of our limb pose-estimation task is to link subsequent joints for each of the infants' limb, which is done on depth images, individually. First, we identify joint candidates from the output joint-confidence maps using non-maximum suppression, which is an algorithm commonly used in computer vision when redundant candidates are present~\cite{hosang2017learning}. 
Once joint candidates are identified, they are linked exploiting the joint-connection confidence maps. In particular, we use a bipartite matching approach, which consists of: (i) computing the integral value along the line connected two candidates on the joint-connection confidence map and (ii) choosing the two winning candidates as those guaranteeing the highest integral value. 

\begin{figure}[tbp]
    \centering
    \begin{subfigure}[b]{0.45\textwidth}
        \centering
        \includegraphics[width = \textwidth]{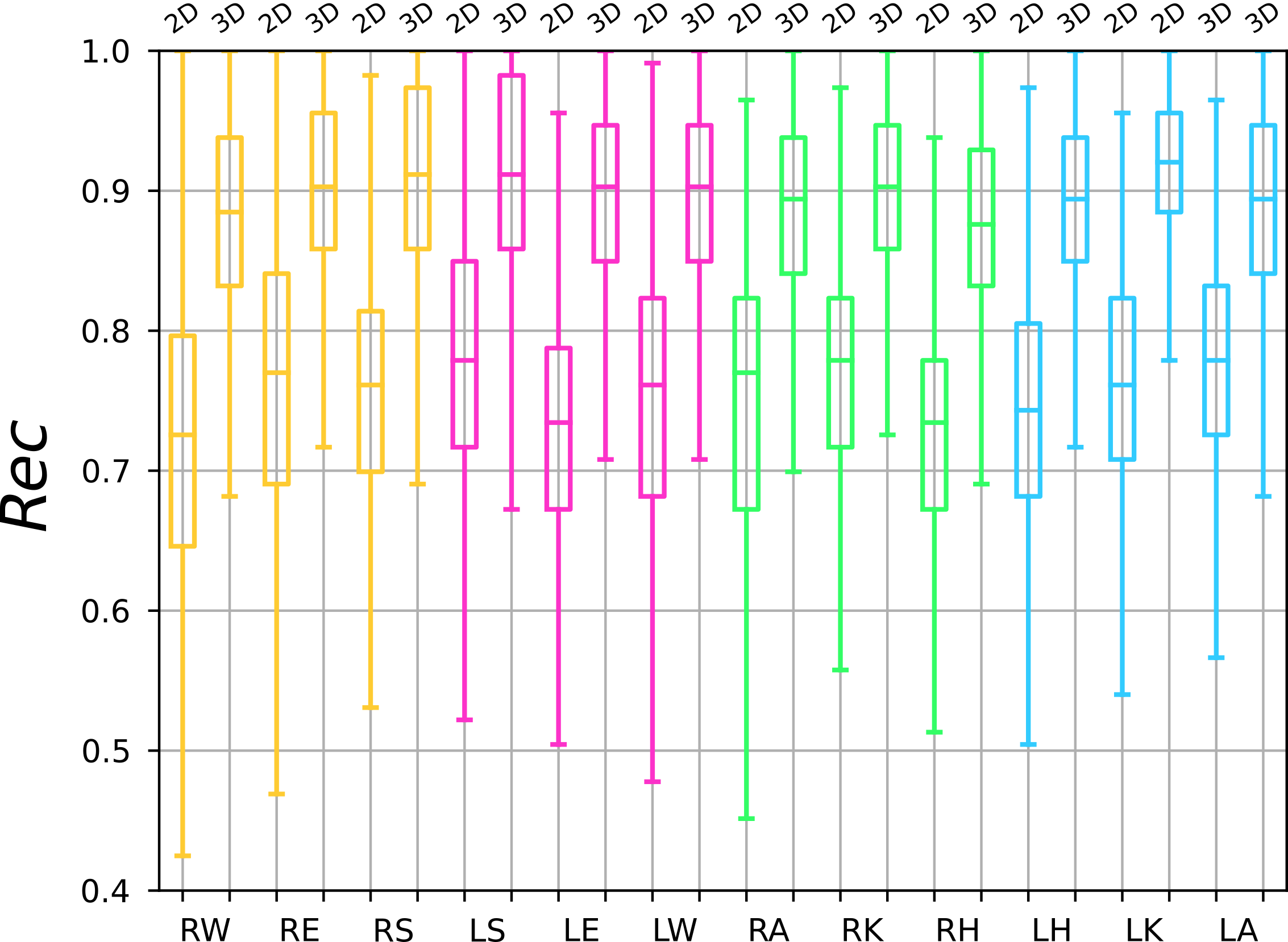}
        \caption{Joint $Rec$}
    \end{subfigure}%
    \\
    \begin{subfigure}[b]{0.45\textwidth}
        \centering
        \includegraphics[width = \textwidth]{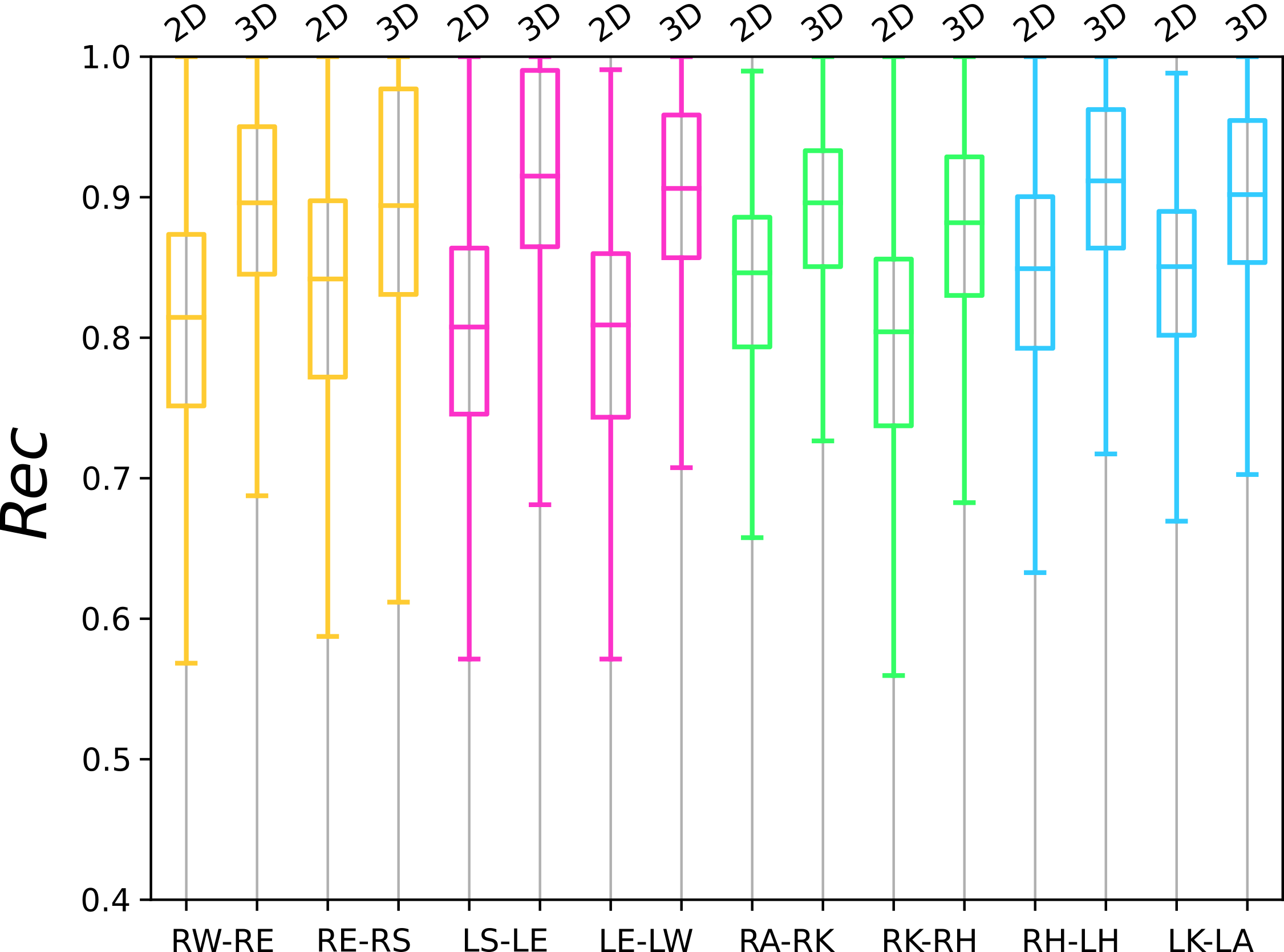}
       \caption{Joint-connetction $Rec$}
    \end{subfigure}
    \caption{\label{fig:dsc_box}Boxplots of the recall ($Rec$) for (a) joint and (b) joint-connection detection achieved with the proposed 3D framework. Results of the 2D framework are shown for comparison, too. For colors and acronyms, refer to the joint model in Fig.~\ref{fig:joint_model}.}
\end{figure}

\section{Evaluation}
\label{sec:exp}

\subsection{Dataset}

Our babyPose dataset consisted of 16 depth videos of 16 preterm infants that were acquired in the NICU of the G. Salesi Hospital in Ancona, Italy. Demographic data for the babyPose dataset are shown in Table \ref{babyPoseD}. The babyPose dataset presents high variability in terms of gestational age (mean=31.87 $\pm$ 3.77), weight (mean = 2021 $\pm$ 790), and length (mean= 44.13 $\pm$ 4.12). Such variability poses further challenges to the problem of pose estimation.

The infants were identified by clinicians in the NICU among those who were spontaneously breathing. 
%
%
Written informed consent was obtained from the infant's legal guardian.

Video-acquisition setup, which is shown in Fig. \ref{fig:system_installation},  was designed to not hinder healthcare operators in their work activities.
The 16 video recordings (length = 180s) were acquired for every infant using the Astra Mini S - Orbbec~\textregistered,  with a frame rate of 30 frames per second and image size of 640x480 pixels. No frame selection was performed (i.e., all frames were used for the analysis).

Joint annotation was performed under the supervision of our clinical partners using a custom-built annotation tool, publicly available online\footnote{\url{https://github.com/roccopietrini/pyPointAnnotator}}.

For each video, the annotation was manually obtained every 5 frames, until 1000 frames per infant were annotated. 
In accordance with our clinical partners, performing manual annotation every 5 frames may be considered a good compromise considering the average preterm infants' movement rate \cite{fallang2003kinematic}.
Then, these 1000 frames were split into training and testing data: 750 frames were used for training purpose and the remaining ones (250 frames) to test the network; resulting in a training set of 12000 samples (16 infants x 750 frames) and a testing set of 4000 samples (16 infants x 250 frames). From the 12000 training samples, we kept 200 frames for each infant as validation set, for a total of 3200 frames.  
%
\begin{table*}[tbp]
\centering
\caption{\label{tab:res_rmsd} Limb-pose estimation performance in terms of median root mean square distance ($RMSD$), with interquartile range in brackets, computed with respect to the ground-truth pose. The $RMSD$ is reported for each limb, separately. Results are reported for the 2D and 3D framework, as well as for the 3D detection-only, 3D regression-only and state-of-the-art architectures.}
{\renewcommand\arraystretch{1.2} 
\begin{tabular}{c|c|c|c|c}
& Right arm & Left arm   & Right leg  & Left leg \\ 
\hline
& \multicolumn{4}{c}{$RMSD$}\\
\hline
2D& 11.73 (3.58) & 10.54 (4.97) & 11.03 (5.78) & 11.50 (4.21) \\
Detection-only network & 15.09 (3.80) & 15.60 (3.87) & 15.09 (3.41) & 14.91 (3.49) \\
Regression-only network & 12.39 (2.18) & 11.73 (3.25) & 11.95 (4.60) & 12.17 (2.47) \\
Stacked Hourglass & 13.01 (4.12) & 11.95 (4.60) & 11.27 (5.32) & 11.95 (3.58) \\
Convolutional Pose Machine & 12.17 (4.52) & 11.73 (3.65) & 11.27 (4.61) & 11.95 (3.44)\\
3D& 9.76 (4.60) & 9.29 (5.89) & 8.90 (5.64) & 9.20 (3.99)  \\
\end{tabular}}
\end{table*}

Figure \ref{fig:challenges} shows some of the challenges in the dataset, such as varying infant-camera distance (due to the motility of the acquisition setup), different number of visible joints (due to partial or total limb occlusion) and presence of homogeneous areas with similar or at least continuous intensity values, due to the use of the depth video.

\subsection{Training settings}
\label{sec:train_sett}
All frames were resized to 128x96 pixels in order to smooth noise and reduce both training time and  memory requirements. Mean intensity was removed from each frame.
To build the ground-truth masks, we selected $r_d$ equal to 6 pixels, as to completely overlay the joints. The $W_s$ was set to 2 for training and 0 for testing, while $W_d$ was set to 3. This way, a temporal clip was 0.5s long.  

For training the detection and regression network, we set an initial learning rate of 0.01 with a learning decay of 10\% every 10 epochs, and a momentum of 0.98. We used a batch size of 8 and set a number of epochs equal to 100.
We selected the best model as the one that maximized the detection accuracy and minimized the mean absolute error on the validation set, for the detection and regression network, respectively.

All our analyses were performed using Keras\footnote{\url{https://keras.io/}} on a Nvidia GeForce GTX 1050 Ti/PCIe/SSE2.

\begin{figure}[tbp]
\centering	
\includegraphics[width=.45\textwidth]{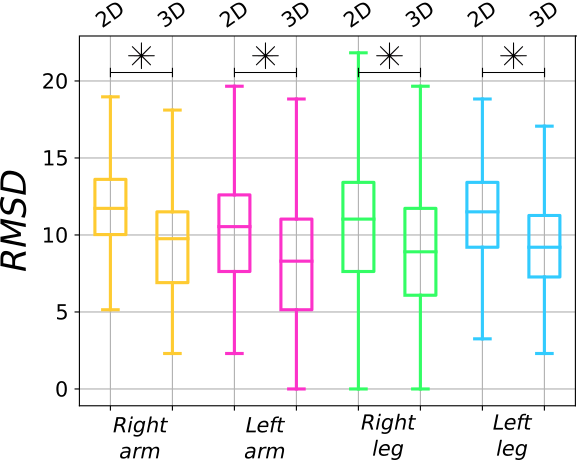}
\caption{Boxplots of the root mean squared distance ($RMSD$) computed for the four limbs separately. Boxplots are shown for the 2D and 3D framework. Asterisks highlight significant differences.}
\label{fig:pose_boxplot}
\end{figure}



\subsection{Ablation study and comparison with the state of the art}
\label{sec:ablation}

We compared the performance of the proposed spatio-temporal features with that of spatial features alone.  We chose  the closest work with respect to ours (i.e., ~\cite{moccia2019preterm}), which is inspired by \cite{cao2017realtime} and uses the same architectures presented in Table~\ref{tab:archi_det} and Table~\ref{tab:archi_reg}, but with 2D spatial convolution. 

We also compared the proposed approach with the Stacked Hourglass \cite{newell2016stacked} and Convolutional Pose Machine \cite{wei2016convolutional}, which are among the most successful and well-known approaches for human pose estimation.
For these comparisons, we modified the corresponding architectures\footnote{\url{https://github.com/yuanyuanli85/Stacked_Hourglass_Network_Keras/blob/master/src/net/hg_blocks.py}}\textsuperscript{,}\footnote{\url{https://github.com/namedBen/Convolutional-Pose-Machines-Pytorch/blob/master/train_val/cpm_model.py}}, originally designed for RGB images, to allow depth-image processing.
For all these architectures, we implemented the same training settings described in Sec.~\ref{sec:train_sett}.

For the ablation study, inspired by \cite{du2018articulated}, we compared the performance of the proposed framework with the detection-only and regression-only architectures. Both were implemented in a spatio-temporal fashion (i.e., with 3D convolution).
For   the   detection-only  model, the affinity maps were used to directly estimate limb pose with the bipartite-matching strategy described in Sec.~\ref{sec:bipartile}.
The regression-only  model was fed with the depth clips and trained with the confidence-map ground truth. The output was then used to estimate joint pose with bipartite matching.


\subsection{Performance metrics}

To measure the performance of the detection network, as suggested in \cite{colleoni2019deep}, we computed the Dice similarity coefficient ($DSC$) and recall ($Rec$), which are defined as:

\begin{equation}
DSC = \frac{2 \times TP}{2 \times TP + FP + FN}
\end{equation}
\begin{equation}
Rec = \frac{TP}{TP+FN}
\end{equation}
where $TP$ and $FP$ are the true joint and background pixels detected as joints, respectively, while $FN$ refers to joint pixels that are detected as background. The same applied to joint connections.  

To evaluate the overall pose estimation, we computed the root mean square distance ($RMSD$) [pixels] for each infants' limb.
For both the detection and regression network, we measured the testing time.

Two-sided t-test with significance level ($\alpha$) = 0.05 was used to evaluate if significative differences were present between the 2D and 3D framework in estimating limb pose.


\begin{figure}[tbp]
\centering	
\includegraphics[width=.5\textwidth]{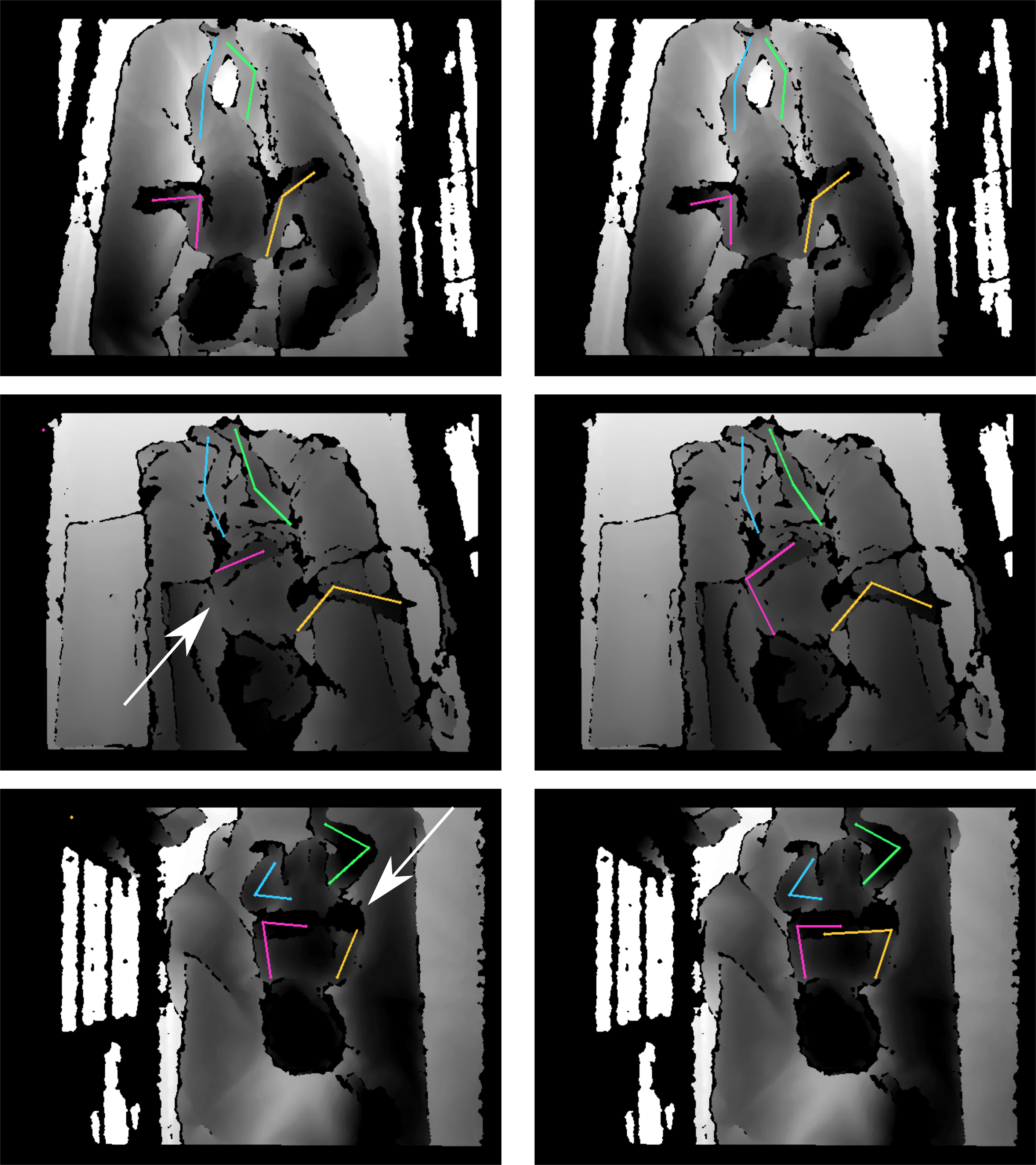}
\caption{Sample qualitative results for pose estimation obtained with the 2D (left) and 3D (right) framework. White arrows highlight estimation errors, mainly due to homogeneous image intensity. 
}
\label{fig:qualitative_analysis_2D_3D}
\end{figure}

\begin{figure*}[tbp]
\centering	
\includegraphics[width=\textwidth]{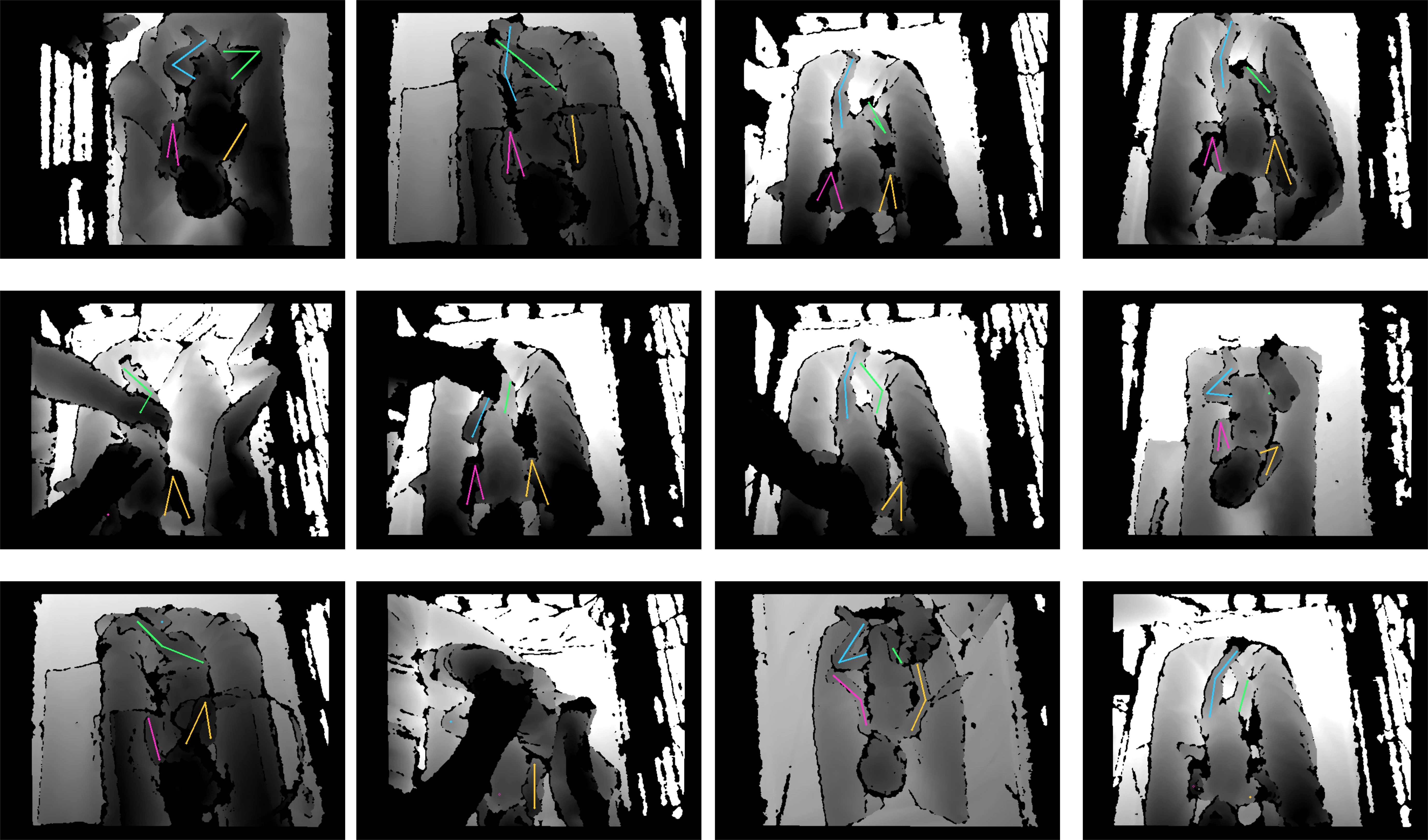}
\caption{Sample qualitative results for challenging cases. 
First row: one joint was not detected due to auto-occlusion (from left to right: right shoulder, right shoulder, right hip, right hip). Second row: one or more joints were not detected due to external occlusion (from left to right: joint of the left limbs, right ankle, left arm - due to healthcare operator hand presence, and right knee and ankle - due to plaster). Last row: image noise and intensity inhomogeneities prevented joint detection.}
\label{fig:qualitative_analysis}
\end{figure*}

\section{Results}
\label{sec:res}

The descriptive statistics of $Rec$ and $DSC$ for the detection CNN are reported in Table~\ref{tab:res_joint_detection}. Figure \ref{fig:dsc_box}a shows the $Rec$ boxplots for joints. Results are also shown for the corresponding 2D implementation.
The highest median $DSC$ (0.94, inter-quartile range (IQR) = 0.05) among all joints was obtained with the 3D CNN. The same was observed for the $Rec$, with a median value among all joints of 0.90, and IQR of 0.09.
Note that, in the case yielding the least accurate result, which corresponds to the RH joint, the $Rec$ still achieved 0.88, whereas for the 2D detection network the lowest $Rec$ was 0.73.
The same behaviour (Table \ref{tab:res_joint_connection} and Fig. \ref{fig:dsc_box}b) was observed when considering the joint-connection detection performance, with median $DSC$ = 0.93 (IQR = 0.06) and median $Rec$ = 0.90 (IQR = 0.11) among all connections.
%
%


The performance  comparison in terms of $RMSD$ of the different  models  presented in Sec. \ref{sec:ablation} is summarized in Table \ref{tab:res_rmsd}. The highest performance (i.e., the lowest $RMSD$) was achieved by the 3D framework, with a median value of 9.06 pixels (IQR = 5.12) among the four limbs. The best performance was achieved for the right leg (median = 8.90 pixels, IQR = 5.64 pixels). The overall computational time for our 3D framework was 0.06s per image on average.
The 2D framework always showed lower performance, with the best and worst $RMSD$ equal to 10.54 (left arm) and 11.73 (right arm) pixels, respectively (median among the four limbs = 11.27 with IQR = 4.59). The overall statistics are shown in Fig. \ref{fig:pose_boxplot}. The results all differed significantly (p-value $< \alpha$) from those obtained with the 3D framework. %
Stacked Hourglass and Convolutional Pose Machine got a median $RMSD$ of 11.95 and 11.84 pixels.
The detection-only and regression-only networks showed the lowest performance, with a median $RMSD$ equal to 15.09 pixels and 12.06 pixels, respectively. 

In Fig. \ref{fig:qualitative_analysis_2D_3D}, qualitative results for  infants' pose estimation are shown both for the 2D framework (on the left side) and the 3D one (on the right side). The white arrows highlight errors in pose estimation made by the 2D framework. 
Results of the 3D framework for challenging cases are shown in Fig.~\ref{fig:qualitative_analysis}. %
The first row shows samples in which  one joint  was  not detected  due  to  auto-occlusion. Joints were also not detected when external occlusion occurred (second row), due to the interaction of the healthcare-operator with the infant or to the presence of plaster. The proposed framework was not able to produce joint estimation also when image noise and intensity inhomogeneities (e.g., due to rapid infants movement) were present (third row). At the same time, however, other joints in the image were correctly estimated thanks to the joint-map parallel processing.

\section{Discussion}
\label{sec:disc}

Monitoring preterm infants' limb is crucial for assessing infant's health status and early detecting cognitive/motor disorders. 
However, when surveying the clinical literature, we realized that there is a lack of documented quantitative parameters on the topic. This is mainly due to the drawbacks of current monitoring techniques, which rely on qualitative visual judgment of clinicians at the crib side in NICUs.
A possible, straightforward, solution may be to exploit contact sensors (such as accelerometers). Nonetheless, in NICUs, using additional hardware may contribute significantly
to infants' stress, discomfort and pain and, from the healthcare operators' point of view, may hinder the actual clinical practice.
To overcome all these issues, researchers seek for new reliable and unobtrusive monitoring alternatives, which are mostly based on video analysis.
With this paper, we proposed a novel framework for non-invasive monitoring of preterm infants' limbs through providing an innovative approach for limb-pose estimation from spatio-temporal features extracted from depth videos. We decided to exploit depth videos (over approaches based on RGB videos) to accomplish considerations relevant to privacy protection.
The deep learning framework was validated on a
dataset of 16 preterm infants, whose video recordings, acquired in the actual clinical practice, presented several challenges such as: presence of homogeneous areas with similar or at least continuous
intensity, self- or external occlusions and different pose of the camera with respect to the infants.

%

%
The proposed 3D detection network achieved encouraging results as shown in Fig. \ref{fig:dsc_box} and reported in Table \ref{tab:res_joint_detection}, with a median $DSC$ of 0.94 and 0.93 for joint and  joint-connection, respectively, overcoming our previous approach based on spatial features only \cite{moccia2019preterm}. The network performed comparably when detecting all joints and joint-connection as shown by the IQRs in Table \ref{tab:res_joint_detection}, reflecting the CNN ability of processing in parallel the different joint and joint-connection affinity maps.
%

The 3D framework achieved improved performance (Table~\ref{tab:res_rmsd}) in estimating infants' pose for all limbs (median $RMSD$ $=$ 9.06 pixels) when compared with our previous 2D approach (median $RMSD$ $=$ 11.27 pixels). 
These results suggest that exploiting temporal information improved network generalization ability even in presence of  intensity homogeneity and noisy background, typical of depth images. These considerations are visible in Fig. \ref{fig:qualitative_analysis_2D_3D}, where the 2D framework failed in detecting joints that lay in portions of the image with homogeneous intensity. 

Predictions of the pose estimation were computed also for the detection-
(median $RMSD$ $=$ 15.09 pixels) and the regression-only networks (median $RMSD$ $=$ 12.06 pixels).
Despite the complexity of regressing joint and joint-connection confidence maps from depth image clips only, the regression-only network achieved better results when compared to the detection-only network. 
The lower performance of the detection-only network may be due to the complexity in localizing joint candidates from ground-truth binary masks, where all pixels have the same weight (Fig. \ref{fig:ground_truth_detection}). 
It is worth noting that spatio-temporal features were tested for a detection-only task in \cite{colleoni2019deep} (even though for surgical instrument joints in laparoscopic video). Here, however, we moved forward to test joint estimation by combining the detection network with bipartite matching, and comparing the achieved results with the full 3D detection+regression framework.
Despite the integration of the temporal information, both  the detection-only and regression-only network achieved lower outcomes with respect to the full 2D framework.
Hence, the regression-only  model  was  barely capable  of predicting  the location of joints without any guidance. Regression is  empirically too  localized  (i.e., it  supports small  spatial context)  and the  process  of  regressing  from  original  input image to joint location directly is challenging. 
By combining detection  and  regression, the  detection  module acted as structural guidance for the regression module by providing spatial contextual information between joints, and facilitating the joints localization.  

Stacked Hourglass and Convolutional Pose Machine achieved lower performance when compared to our 3D framework. This might be attributed to the fact that both Stacked Hourglass and Convolutional Pose Machine are designed to process spatial features only.  
Nonetheless, the 2D framework, which also works with spatial feature only, overcame both Stacked Hourglass and Convolutional Pose Machine. This result seems to confirm that the rough detection of limb joints by the detection network facilitates the regression network in regressing joint position accurately, as highlighted in \cite{cao2017realtime}.
In fact, Stacked Hourglass and Convolutional Pose Machine achieved better $RMSD$ values when compared to the regression-only network. Hence, the benefits brought by the introduction of 3D kernels in the regression-only network are counterbalanced by the multi-scale nature of the state-of-the-art networks, which capture both global and local information.

\begin{figure}[tbp!]
    \centering
    \includegraphics[width=.5\textwidth]{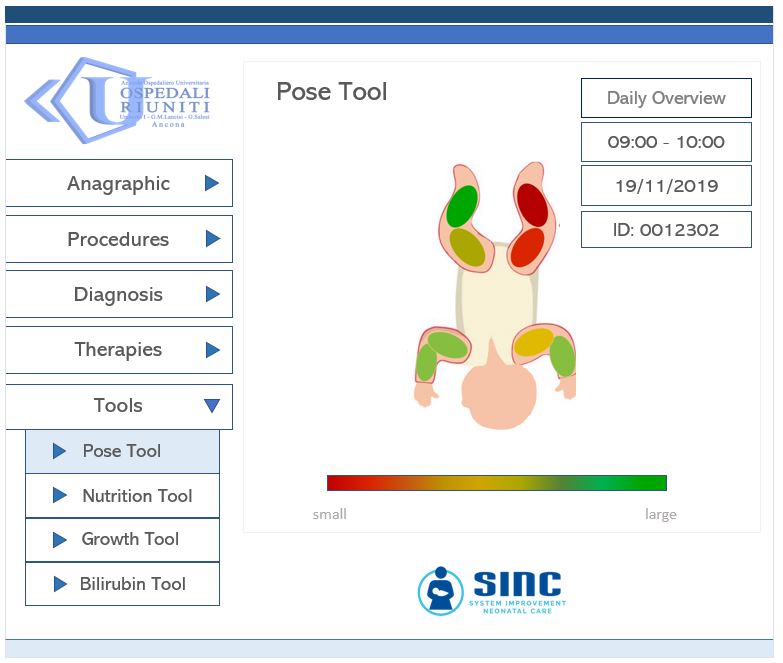}
    \caption{Graphical user interface of the Pose Tool, which codifies with a color code the standard deviation of the joint angles in time.}
    \label{fig:my_GUI}
\end{figure}

A straightforward limitation of this work may be seen in the estimation of occluded joints (both in case of auto and external occlusion), as shown in Fig. \ref{fig:qualitative_analysis} (first and second rows). At the same time, our two-branch architecture with multiple maps allowed to detect the other (not-occluded) joints in the image. 
%
%
This issue could be attenuated with recent strategies proposed in the literature for long-term tracking  \cite{penza2018long} and confidence estimation~\cite{moccia2018uncertainty}.
Modeling infant's limbs through anthropometric measures (such as limb length - already acquired in the actual clinical practice) could also help in attenuating the occlusion issue. This would probably also make our 3D framework able to tackle noisy image portions, which may be present due to sudden movement of infants or healthcare operators (Fig. \ref{fig:qualitative_analysis}, last row).
We also recognize that a limitation of the proposed
work could be seen is the relatively limited number of testing frames (4000), which is due to the lack of available annotated
dataset online. To attenuate this issue, we released the data we
collected for further use in the community. 
%
%

As future work, to support clinicians in the actual clinical practice, we plan to develop a tool based on limb-pose estimation (the Pose Tool) to be integrated within the electronic medical-record software currently in use in the NICU of the G. Salesi Hospital. Starting from the limb-pose estimation, the joint angles can be computed (e.g. according to \cite{khan2018detection}), offering useful hints for infants' monitoring \cite{sweeney2002musculoskeletal}.
Figure \ref{fig:my_GUI} shows the graphical user interface of the Pose Tool, which codifies with a color code the standard deviation of the joint angles in time.
Natural extensions of the proposed work deal with the inclusion of the limb-pose estimation within other computed-assisted algorithms for diagnostic support, e.g., to classify abnormal limb movements. 
The proposed acquisition setup could also be integrated with recent video-based monitoring systems for respiratory rate analysis \cite{pereira2018noncontact}.


\section{Conclusion}
\label{sec:concl}
In  this  paper,  we  proposed  a  framework for preterm infants' limb-pose estimation from depth images based on spatio-temporal features.
Our results, achieved by testing a new contribution dataset (which is also the first in the field),  suggest  that  spatio-temporal
features can be successfully exploited to increase pose-estimation
performance with respect to 2D models based on single-frame (spatial only)
information.

In conclusion, our solution moves us towards a better framework for preterm infants' movement
understanding  and  can  lead  to  applications  in computer-assisted diagnosis.
Moreover, by making our  dataset fully available, we believe we will stimulate researches in the field, encouraging and promoting the clinical translation of preterm infants' monitoring systems for timely diagnosis and 
prompt treatment.

 page
\ifCLASSOPTIONcaptionsoff
  \newpage
\fi

\subsection*{Compliance with ethical standards}

\subsection*{Disclosures}
The authors have no conflict of interest to disclose.
\subsection*{Ethical standards} 
The procedures followed were in accordance with the ethical standards of the responsible committee on human experimentation (institutional and national) and with the Helsinki Declaration of 1975, as revised in 2000. 
This article followed the Ethics Guidelines for Trustworthy Artificial Intelligence, recently published by the European Commission\footnote{\url{https://ec.europa.eu/futurium/en/ai-alliance-consultation}}.



\end{document}